\documentclass[10pt,twocolumn,letterpaper]{article}

\usepackage{3dv}
\usepackage{times}
\usepackage{epsfig}
\usepackage{graphicx}
\usepackage{amsmath}
\usepackage{amssymb}
\usepackage{enumitem}
\usepackage{mathtools}
\DeclareMathOperator*{\argmin}{\arg\!\min}
\DeclareGraphicsExtensions{.pdf}

\usepackage[pagebackref=true,breaklinks=true,letterpaper=true,colorlinks,bookmarks=false]{hyperref}

\threedvfinalcopy 

\def\threedvPaperID{19} 

\ifthreedvfinal\fi
\begin{document}

\title{DepthSynth: Real-Time Realistic Synthetic Data Generation from CAD Models for 2.5D Recognition}

\author{
Benjamin Planche$^{1}$, Ziyan Wu$^{2}$, Kai Ma$^{3}$, Shanhui Sun$^{3}$, Stefan Kluckner$^{4}$, Oliver Lehmann$^{3}$,\\ Terrence Chen$^{3}$, Andreas Hutter$^{1}$, Sergey Zakharov$^{1}$, Harald Kosch$^{5}$, Jan Ernst$^{2}$\\
$^{1}$Siemens Corporate Technology, Germany\\
{\tt\small \{benjamin.planche, andreas.hutter, sergey.zakharov\}@siemens.com}
\and
$^{2}$Siemens Corporate Technology, USA\\
{\tt\small \{ziyan.wu, jan.ernst\}@siemens.com}
\and
$^{3}$Siemens Healthineers, USA\\
{\tt\small \{kai.ma, shanhui.sun, oliver.lehmann, terrence.chen\}@siemens.com}
\and
$^{4}$Siemens Mobility, Germany\\
{\tt\small stefan.kluckner@siemens.com}
\and
$^{5}$University of Passau, Germany\\
{\tt\small harald.kosch@uni-passau.de}
}

\maketitle

\begin{abstract}
Recent progress in computer vision has been dominated by deep neural networks trained over large amounts of labeled data. Collecting such datasets is however a tedious, often impossible task; hence a surge in approaches relying solely on synthetic data for their training. For depth images however, discrepancies with real scans still noticeably affect the end performance. We thus propose an end-to-end framework which simulates the whole mechanism of these devices, generating realistic depth data from 3D models by comprehensively modeling vital factors e.g. sensor noise, material reflectance, surface geometry. Not only does our solution cover a wider range of sensors and achieve more realistic results than previous methods, assessed through extended evaluation, but we go further by measuring the impact on the training of neural networks for various recognition tasks; demonstrating how our pipeline seamlessly integrates such architectures and consistently enhances their performance.
\end{abstract}

\section{Introduction}

\begin{figure}[t]
  \centering
  \includegraphics[width=1\linewidth]{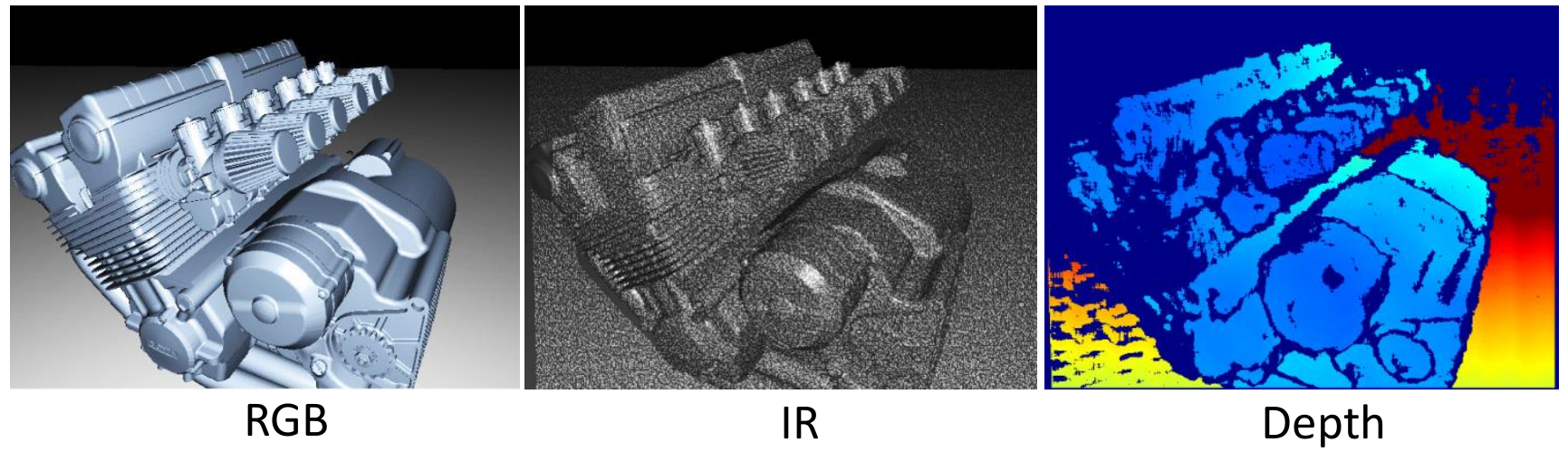}
  \caption{\textbf{Synthetic sample generated by the proposed pipeline}}
  \label{fig:teaser}  
\end{figure}

Understanding the 3D shape or spatial layout of a real-world 
object captured in a 2D image has been a classic computer vision problem for decades~\cite{pr97:sr,tpami99:sfs,cgf03:vsr}. However, with the advent of low-cost 
depth sensors, specifically structured light cameras~\cite{iccv05:sls} 
\eg Microsoft Kinect, Intel RealSense, its focus has seen a substantial paradigm shift. What in the past 
revolved around interpretation of raw pixels in 2D projections has now become 
the analysis of real-valued depth (2.5D) data. This has drastically increased the scope of practical applications ranging from 
recovering 3D geometry of complex surfaces~\cite{ismar11:kf,
puc14:cgar} to real-time recognition of human actions~\cite{cvpr11:hpr}, inspiring research in automatic 
object detection \cite{nips12:d3d,cviu15:synt,eccv12:issi,cvpr15:sund}, 
classification \cite{icra11:lmvd,iros15:vn,cviu15:synt,icra14:bb,nips12:crdl,cvpr15:sund} and 
pose estimation \cite{nips12:d3d,iccv13:fpe,icra09:l3d}.\\
\indent While real data is commonly used for comparison and training, a large number of these recent studies decompose the 
problems to matching acquired depth images of real-world objects to synthetic ones rendered from a database of pre-existing 3D models~\cite{nips12:d3d,cviu15:synt,nips12:crdl,iccv13:fpe,
cvpr15:sund,icra14:bb,carlucci2016deep}. 
With no theoretical upper bound on obtaining synthetic images to either train complex models for classification~\cite{iccv15:mvcnn,iros15:vn,cvpr15:3dsp,
icra09:l3d} or fill large databases for retrieval tasks~\cite{accv10:els,eccv14:rtf}, research continues to gain impetus in this direction.\\ 
\indent Despite the simplicity of the above flavor of approaches, their performance is often restrained by the lack of {\em realism} (discrepancy with real data) or {\em variability} (limited configurability) of their rendering process. As a workaround, some approaches fine-tune their systems on a small set of real scans~\cite{Wohlhart15}; but in many cases, access to real data is too scarce to bridge the discrepancy gap. Other methods try instead to post-process the real images to clear some of their noise, making them more similar to synthetic data but losing details in the process~\cite{cvpr15:3dsp} which can be crucial for tasks such as pose estimation or fine-grained classification.

\begin{figure}[t]
  \centering
  \includegraphics[width=1\linewidth]{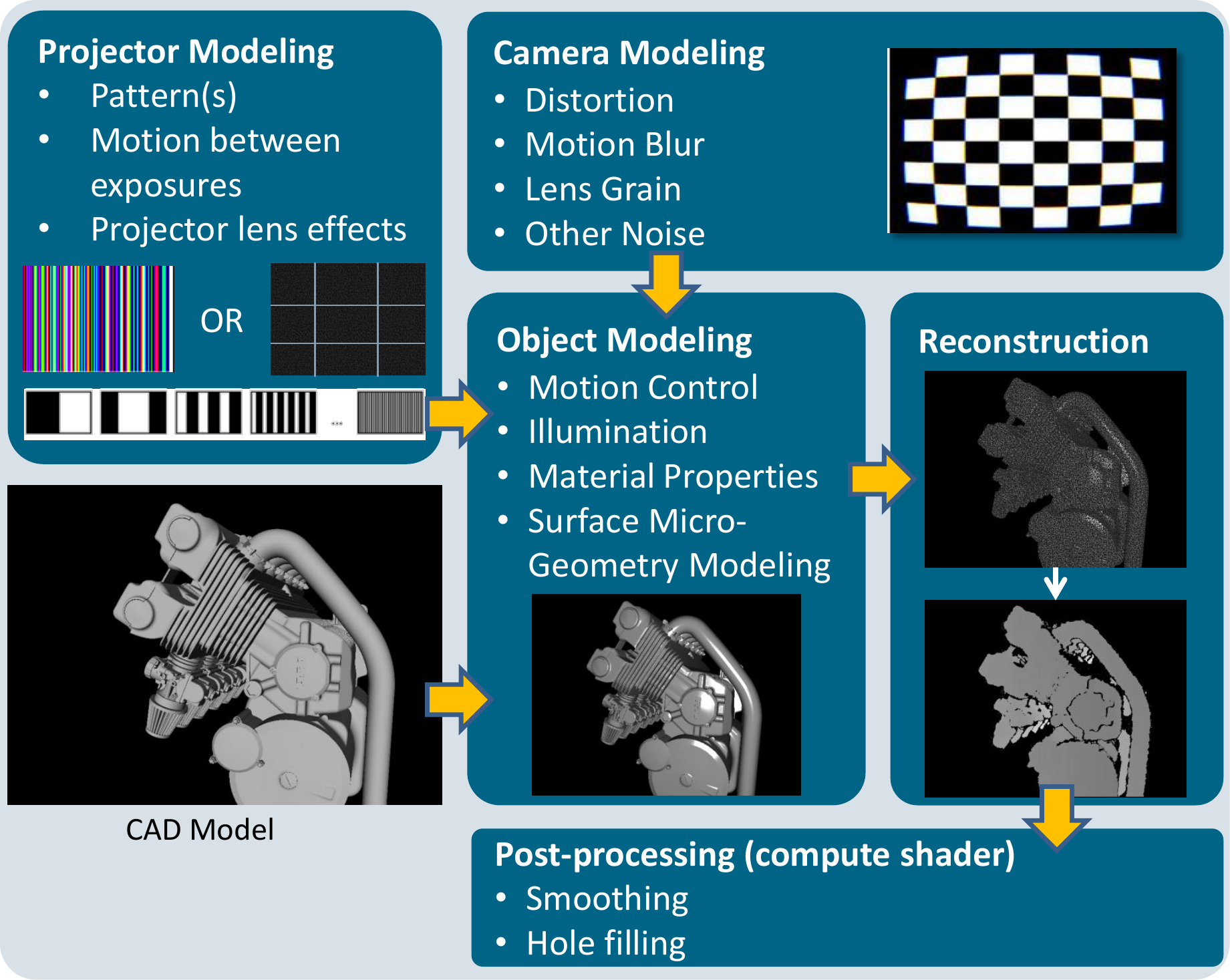}

  \caption{\textbf{Representation of \textit{DepthSynth} pipeline.}}
  \label{fig:pipeline}  
\end{figure}

\indent A practical approach to 
address this problem is thus to generate more data, and in such a way that they mimic captured ones. This is however a non-trivial problem, as it is extremely 
difficult to exhaustively enumerate all physical variations of a 
given object---including surface geometry, reflectance, deformation, 
etc. Addressing those challenges in this paper, our key contributions are as follows: (a) we introduce \textit{DepthSynth}, 
an end-to-end pipeline to synthetically generate depth images from 3D 
models by virtually and comprehensively reproducing the sensors mechanisms (Figure~\ref{fig:pipeline}), replicating 
realistic scenarios and thereby facilitating 
robust 2.5D applications, regardless the ulterior choice of algorithm or feature space; (b) we systematically evaluate and compare the quality of the resulting images with theoretical models and other modern simulation methods; (c) we demonstrate the effectiveness and flexibility of our tool by pairing it to a state-of-the-art method for two recognition tasks.\\
\indent The rest of the paper is organized as follows. In Section~\ref{sec:rw}, we 
provide a survey of pertinent work to the interested reader. Next, in Section
~\ref{sec:mth}, we introduce our framework, detailing each step. In Section~\ref{sec:exp}, we elaborate on our experimental protocol; first comparing the sensing errors induced by our tool to experimental data and theoretical models; then demonstrating the usefulness of our method by applying it to the pose estimation and classification tasks used as examples. We finally conclude with insightful discussions in Section~\ref{sec:cnc}.

\begin{figure}[t]
  \centering
  \includegraphics[width=1\linewidth]{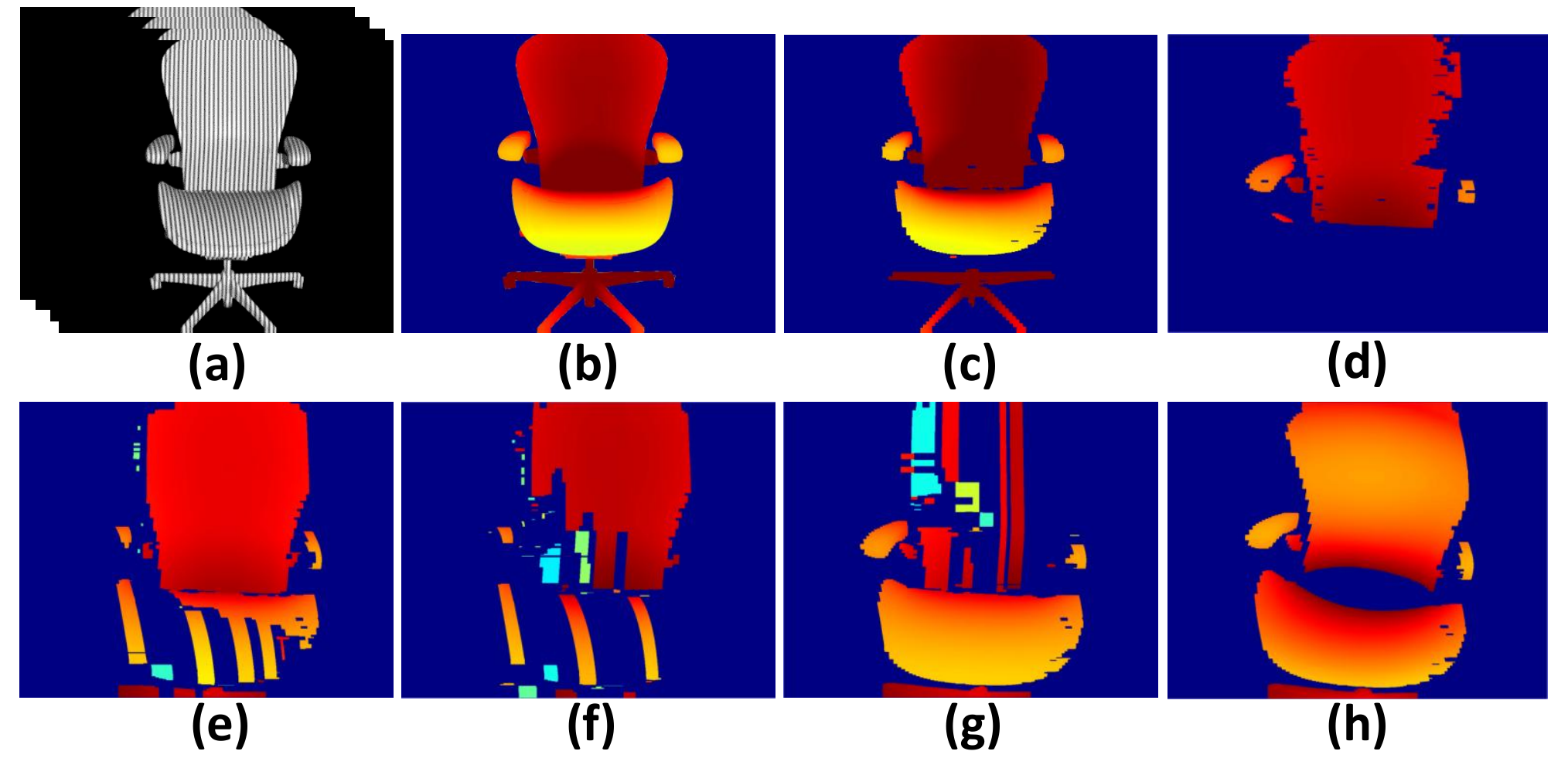}
  \caption{\textbf{Examples of data generated, simulating a multi-shot depth sensor:} (a) Rendering of projected patterns under realistic lighting and surface materials; (b) Ideal depth data; (c) \textit{DepthSynth} generated data without motion or ambient light; (d) with strong ambient light; (e) with motion between exposures (5 cm/s constant speed); (f) with motion between exposures (10 cm/s constant speed); (g) with vibration (2 cm amplitude); (h) with rolling shutter effect (10 cm/s constant speed).}
  \label{fig:multishot}  
\end{figure}

\section{Related Work}
\label{sec:rw}
With the popular advocacy of 2.5D/3D sensor for vision applications, depth information is the support of active research within computer vision. We emphasize on recent approaches which employ synthetic scans, and present previous methods to generate such 2.5D data from CAD models.

\par\noindent
\textbf{Depth-based Methods and Synthetic Data}
Crafting features to efficiently detect objects, discriminate them, evaluate their poses, etc. has long been a tedious task for computer vision researchers. With the rise of machine learning algorithms, these existing models have been complemented \cite{icra09:l3d,pr97:sr,bmvc10:3dcad}, before being almost fully replaced by statistically-learned representations. Multiple recent approaches based on deep convolutional neural networks unequivocally outshone previous methods \cite{cvpr15:3dsp, iros15:vn,iccv15:mvcnn,cvpr15:3dsp,ax15:rcnn}, taking advantage of growing image datasets (such as \textit{ImageNet} \cite{deng2009imagenet}) for their extensive training. As a matter of fact, collecting and accurately labeling large amounts of real data is however an extremely tedious task, especially when 3D poses are considered for ground truth.

In order to tackle this limitation, and concomitantly with the emergence of 3D model databases, renewed efforts  \cite{rematas2014image,ax15:rcnn} were put into the synthetic extension of image or depth scan datasets, by applying various deformations and noise to the original pictures or by rendering images from missing viewpoints. These augmented datasets were then used to train more flexible estimators. Among other deep learning-based methods for class and pose retrieval recently proposed \cite{cvpr15:3dsp, iros15:vn,iccv15:mvcnn}, Wu~\etal \textit{3D Shapenets} \cite{cvpr15:3dsp} and Su~\etal \textit{Render-for-CNN} \cite{ax15:rcnn} methods are two great examples of a second trend: using the \textit{ModelNet} \cite{cvpr15:3dsp} and \textit{ShapeNet} \cite{shapenet2015} 3D model datasets they put together, they let their networks learn features from this purely synthetic data, achieving consistent results in object registration, next-best-view prediction or pose estimation. 

Diving further into the problem of depth-based object classification and pose estimation chosen as illustration in this paper, Wohlhart~\etal \cite{Wohlhart15} recently developed a scalable process addressing a two-degree-of-freedom pose estimation problem. Their approach evaluates the similarity between descriptors learned by a Convolutional Neural Network (CNN) with Euclidean distance, followed by nearest neighbor search. They trained their network with real captured data, but also simplistic synthetic images rendered from 3D models.
In our work, this framework is extended to recognizing 3D pose with six degrees of freedom (6-DOF), and fed only with realistic synthetic images from \textit{DepthSynth}. This way we achieve a significantly higher flexibility and scalability of the system, as well as a more seamless application to real-world use cases.


\par\noindent
\textbf{Synthetic Depth Image Generation}
Early research along this direction involves the work of  ~\cite{trs78:rep,ijcai79:mbvs}, wherein search based on 3D representations are introduced. More recently, Rozantsev~\etal presented a thorough method for generating synthetic images \cite{cviu15:synt}. Instead of focusing on making them look similar to real data for an empirical eye, they worked on a similarity metric based on the features extracted during the machine training.
However, their model is tightly bound to properties impairing regular cameras (e.g. lighting and motion blur), which can't be applied to depth sensors.

Su~\etal worked concurrently on a similar pipeline \cite{ax15:rcnn}, optimizing a synthetic RGB image renderer for the training of CNNs. While working on finding the best compromise between quality and scalability, they notice the ability CNNs have to \textit{cheat} at learning from too simplistic images (e.g. by using the constant lighting to deduce the models poses, or by relying too much on contours for pictures rendered without background, etc.). Their pipeline has thus been divided into three steps: the rendering from 3D models, using random lighting parameters; the alpha composition with background images sampled from the SUN397 dataset \cite{sundatabase}; and randomized cropping. By outperforming state-of-the-art pose estimation methods with their own one trained on synthetic images, they demonstrated the benefits such pipelines can bring to computer vision.

Composed of similar steps as the method above, \textit{DepthSynth} can also be compared to the one by Landau~\etal \cite{landau2015simulating,landau2016}, reproducing the Microsoft Kinect's behavior by simulating the infrared capture and stereo-matching process. Though their latter step inspired our own work, we opted for a less empirical, more exhaustive and generic model for the simulated projection and capture of pattern(s). Similar simulation processes were also developed to reproduce the results of Time-of-Flight (ToF) sensors~\cite{peters2008bistatic,keller2009real}. If this paper mostly focuses on single- or multi-shot structured-light sensors, \textit{DepthSynth}'s genericity allows it to also simulate ToF sensors, using a subset of its operations (discarding the baseline distance within the device, defining a simpler projector with phase shift, etc.). Such a subset is then comparable to the method developed by Keller \etal~\cite{keller2009real}. For the sake of completeness, tools such as \textit{BlenSor}\cite{gschwandtner2011blensor}, or \textit{pcl::simulation}\cite{fallon2012efficient,aldoma2012point} should also be mentioned. However, such simulators were implemented to help testing vision applications, and rely on a more simplistic modeling of the sensors, e.g. ignoring reflectance effects or using fractal noise for approximations.

\section{Methodology}
\label{sec:mth}

Our end-to-end pipeline for low-latency generation of realistic depth images from 3D CAD data covers various types of 3D/2.5D sensors including single-shot/multi-shot structured light sensors, as well as Time-of-Flight 
(ToF) sensors (relatively simpler than structured-light ones to simulate, using a sub-set of the pipeline's components e.g. i.i.d. per-pixel noise based on distance and object surface material, etc.). 
From here, we will mostly focus on single-shot sensors e.g. Microsoft Kinect, Occipital Structure and Xtion Pro Live, given their popularity among the research community. 
This proposed pipeline can be defined as a sequence of procedures directly inspired by the underlying mechanism of the sensors we are simulating; i.e. from pattern projection and capture, followed by pre-processing and depth reconstruction using the acquired image and original pattern, to post-processing; as illustrated in Figure~\ref{fig:pipeline}.


\begin{table}[t!]
\footnotesize
\centering
\caption{\textbf{Comparing \textit{BlenSor}~\cite{gschwandtner2011blensor}, Landau's pipeline~\cite{landau2015simulating,landau2016} and ours on sensor noise types.}}
\label{tab:compare_noise_types}
\begin{tabular}{|l|c|c|c|}
 \hline
Type of Noise  & BlenSor & Landau's & DepthSynth\\
 \hline
Axial and Lateral Noise & Yes &  Yes & Yes\\
Specular Surface  & Yes & No & Yes \\
Non-specular Surface   & No & No & Yes\\
Structural Noise &   No & Partial & Yes\\
Lens Distortion and Effects   & No & No & Yes \\
Quantization Step Noise   & No & Yes & Yes\\
Motion and Rolling Shutter   & No & No & Yes\\
Shadow   & No & Partial & Yes\\
 \hline
\end{tabular}
\end{table}

\subsection{Understanding the Noise Causes}
\label{sec:unc}

To realistically generate synthetic depth data, we need first to understand the causes behind the various kinds of noise one can find in the scans generated by real structured light sensors. 
We thus analyzed the different kinds of noise impairing structured light sensors, and their sources and characteristics. This study highlighted how each step of the sensing process introduces its own artifacts.
During the initial step of projection and capture of the pattern(s), noise can be induced by the lighting and material properties of the surfaces (too low or strong reflection of the pattern), by the composition of the scene (e.g. pattern's density per unit area drops quadratically with increasing distance causing axial noise, non-uniformity at edges causing lateral noise, and objects obstructing the path of the emitter, of the camera or both causing shadow noise), or by the sensor structure itself (structural noise due to its low spatial resolution or the warping of the pattern by the lenses).
Further errors and approximations are then introduced during the block-matching and hole-filling operations---such as structural noise caused by the disparity-to-depth transform, band noise caused by windowing effect during block correlation, or growing step size as depth increases during quantization.

By using the proper rendering parameters and applying the described depth data reconstruction procedure, the proposed synthetic data generation pipeline is able to exhaustively induce the aforementioned types of noise, unlike other state-of-the-art depth data simulation methods, as highlighted by the comparison in Table~\ref{tab:compare_noise_types}.

\subsection{Pattern Projection and Capture}
\label{sec:ppac}

In the first part of the presented pipeline, a simulation platform is used to reproduce the pattern projection and capture mechanism. Thanks to an extensive set of parameters, this platform is able to behave like a large panel of depth sensors. Indeed, any kind of pattern can first be provided as an image asset/spotlight cookie for the projection, in order to adapt to the sensing device one wants to simulate. Moreover, the intrinsic and extrinsic parameters of the camera and projector are configurable.

Our procedure covers both the full calibration of structured light sensors and the reconstruction of their projected pattern with the help of an extra camera. Once the original pattern obtained, our pipeline automatically generates a square version of it (to efficiently use spotlight simulation with cookies, projected patterns need to be padded to a square format for the 3D engine), followed by other different ones later used as reference in the block matching procedure according to the camera resolution.

Once obtained, these parameters can be handed to the 3D platform to initialize the simulation. The 3D models must then be provided, along with their material(s). Even though not all models come with realistic textures, the results quality highly depends on such characteristics---especially their reflectance (physically based rendering model~\cite{pharr2016physically} or bidirectional reflectance distribution function~\cite{schlick1994inexpensive}).

Given a list of viewpoints, the platform will perform each pattern capture and projection, simulating realistic illumination sources and shadows, taking into account surface and material characteristics. Along the object, the 3D scene is thus populated with:

\begin{itemize}[noitemsep,nolistsep]
\item A spot light projector, using the desired high resolution pattern (\eg 2000 px by 2000 px) as light cookie;
\item A camera model, set up with the intrinsic and extrinsic parameters of the real sensor, separated from the projector by the provided baseline distance in the horizontal plan of the simulated device;
\item Optionally additional light sources, to simulate the effect of environmental illuminations;
\item Optionally other 3D models (e.g. ground, occluding objects, etc.), to ornament the scene.
\end{itemize}

These settings and procedure allow our method to reproduce complex realistic effects by manipulating camera movement and exposure; e.g. rolling shutter effect can be simulated by acquiring 1 pixel-line per exposure while the camera is moving, or motion blur by averaging several exposures over movement.

Using rendering components implemented by any recent 3D engine with the aforementioned parameters \eg the virtual light projector provided with the pattern(s) and a virtual camera with the proper optical characteristics, we can simulate the light projection / capture procedures done by the real devices, and obtain a ``virtually captured" image with the chosen resolution, similar to the intermediate output of the devices (\eg IR image of the projected pattern).

\subsection{Pre-processing of Pattern Captures}
\label{sec:ppoii}

This intermediate result, captured in real-time by the virtual camera, is then pre-processed (fed into a \textit{compute shader} layer), in order to get closer to the original quality, impinged by imaging sensor noise. In this module, noise effects are added, including radial and tangential lens distortion, lens scratch and grain, motion blur, and independent and identically distributed random noise.

\subsection{Stereo-matching}
\label{sec:sm}

Relying on the principles of stereo vision, the rendered picture is then matched with its reference pattern, in order to extract the depth information from their disparity map. The emitted pattern and the resulting capture from the sensor are here used as the stereo stimuli, with these two virtual eyes (the projector and the camera) being separated by the \textit{baseline} distance $b$. The depth value $z$ is then a direct function of the disparity $d$ with $z = f \cdot b / d$, where $f$ is the focal length in pixel of the receiver.

The disparity map is computed by applying a block-matching process using small \textit{Sum of Absolute Differences} (SAD) windows to find the correspondences \cite{konolige1998small}, sliding the window along the epipolar line. The function value of SAD for the location $(x,y)$ on the captured image is:
\begin{equation}
F_{SAD}(u,v) = \sum_{j}^{w-1}\sum_{i}^{w-1}|I_s(x+i,y+j) - I_t(x+u+i,y+v+j)|
\end{equation}
\noindent where $w$ is the window size, $I_s$ the image from the camera, and $I_t$ the pattern image. The matched location on the pattern image can be obtained by:
\begin{equation}
(u_m,v_m) = \argmin_{(u,v)}F_{SAD}(u,v) 
\end{equation} 
The disparity value $d$  can be computed by:
\begin{equation}
d = \left\{ \begin{array}{ll}u_m-x & \text{horizontal stereo}\\ v_m-y & \text{vertical stereo} \end{array} \right.
\end{equation}

Based on pixel offsets, each disparity value is an integer. Refinement is done by interpolating between the closest matching block and its neighbors, achieving  a sub-pixel accuracy. Given the direct relation between $z$ and $d$, the possible disparity values are directly bound to the sensor's operational depth range, limiting the search range itself. 

\begin{figure}[t]
  \centering
  \includegraphics[width=1\linewidth]{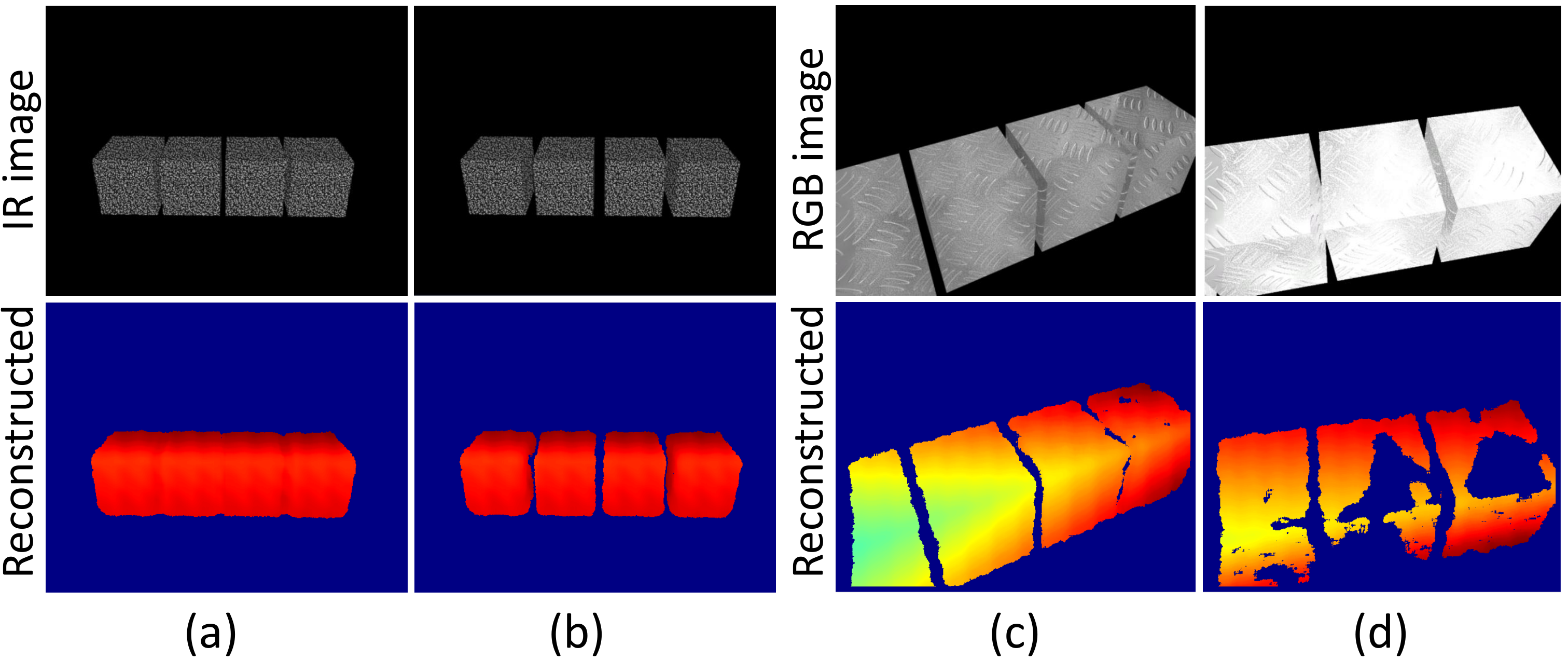}
  \caption{\textbf{Examples of synthetic image pairs for target objects (4 cubes of $\mathbf{0.2 m^3}$) with different placements and surface materials}.
   Cubes in are separated by 5 cm in (a), and by 10 cm in (b, c, d).
   Cubes have fully diffused material in (a, b), $20\%$ reflective material in (c), and $50\%$ reflective material in (d).} 
  \label{fig:sample-single}  
\end{figure}
\begin{figure}[t]
  \centering
  \includegraphics[width=1\linewidth]{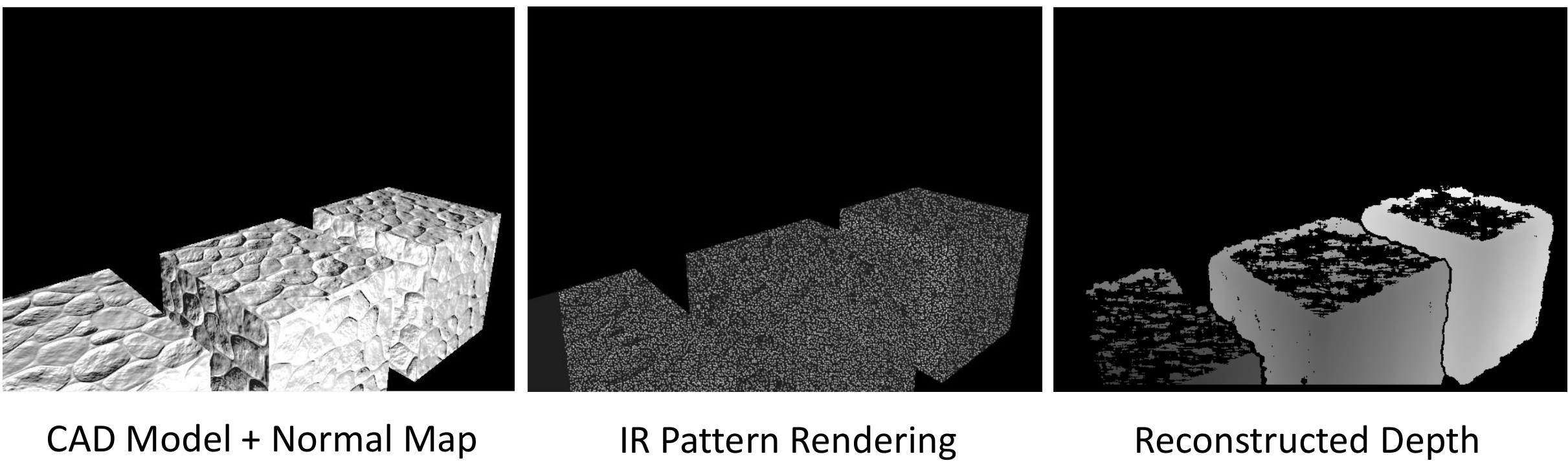}
  \caption{\textbf{Example of the effects of surface conditions on the simulation}, applying a textured normal map to the target objects.}
  \label{fig:surface_condition}  
\end{figure}

\subsection{Post-processing of Depth Scans}
\label{sec:ppods}

Finally, another \textit{compute shader} layer post-processes the depth maps, smoothing and trimming them according to the sensor's specifications. In the case that these specifications are not available, one can obtain reasonable estimation by feeding real images of captured pattern(s) from the sensor through the reconstruction pipeline and derive from the differences between this reconstructed depth image and the one actually output by the sensor. Imitating once more the original systems, a hole-filling step can be performed to reduce the proportion of missing data. 

\begin{figure}[t]
  \centering
  \includegraphics[width=1\linewidth]{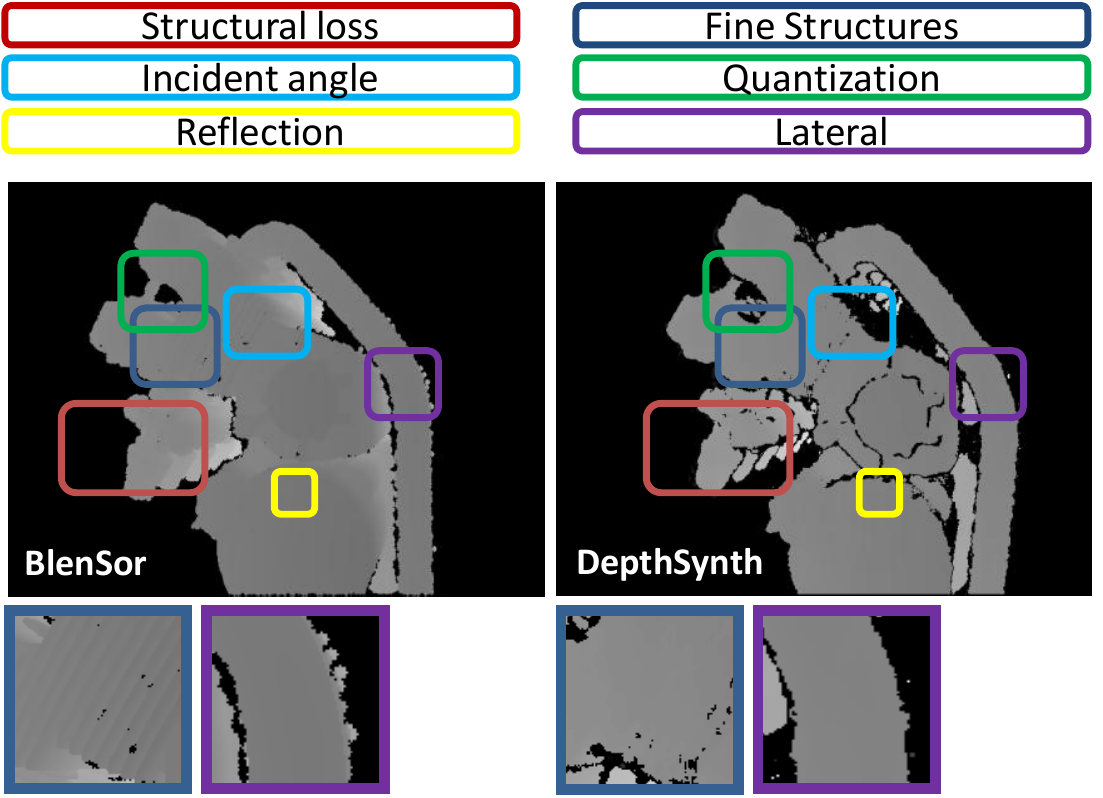}
  \caption{\textbf{Detailed visual comparison with \textit{BlenSor}}~\cite{gschwandtner2011blensor} highlighting the salient differences, based on the noise study presented in Subsection~\ref{sec:unc}.}
  \label{fig:comp_blensor}  
\end{figure}

\begin{figure}[t]
  \centering
  \includegraphics[width=1\linewidth]{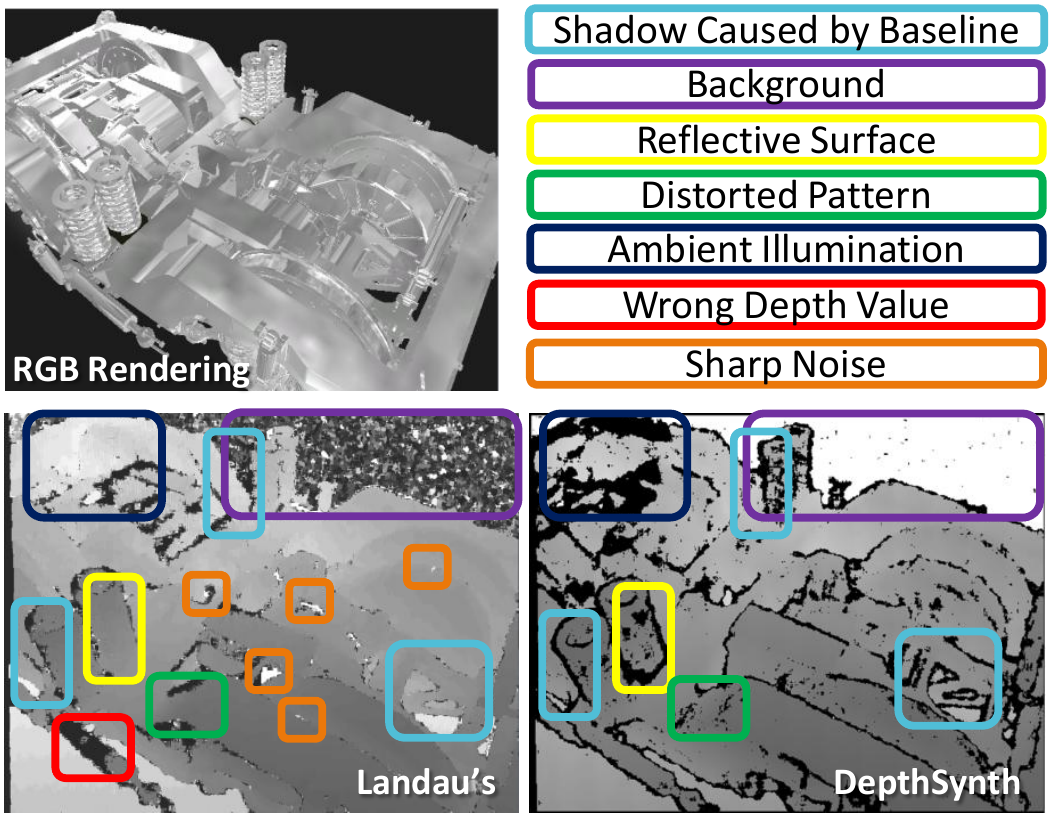}
  \caption{\textbf{Detailed visual comparison with Landau's solution}~\cite{landau2015simulating,landau2016}, based on the noise study presented in Subsection~\ref{sec:unc}.}
  \label{fig:comp_landau}  
\end{figure}

Figures~\ref{fig:sample-single} and \ref{fig:surface_condition} show how \textit{DepthSynth} is able to realistically reproduce the spatial sensitivity of the devices or the impact of surface materials. In the same way, Figure~\ref{fig:multishot} (c)-(h) reveals how the data quality of simulated multi-shot structured light sensors is highly sensitive to motion---an observation in accordance to our expectations.
As highlighted in Figures~\ref{fig:comp_blensor} and~\ref{fig:comp_landau} with the visual comparisons between \textit{DepthSynth} and previous pipelines, the latter ones aren't sensitive to some realistic effects during capture, or preserve fine details which are actually smoothed-out up to the window size in block-matching. Our determination to closely reproduce the whole process performed by the devices paid off in terms of noise quality.






\subsection{Background Blending}
\label{sec:bb}

\begin{figure*}[t]
  \centering
  \includegraphics[width=1\linewidth]{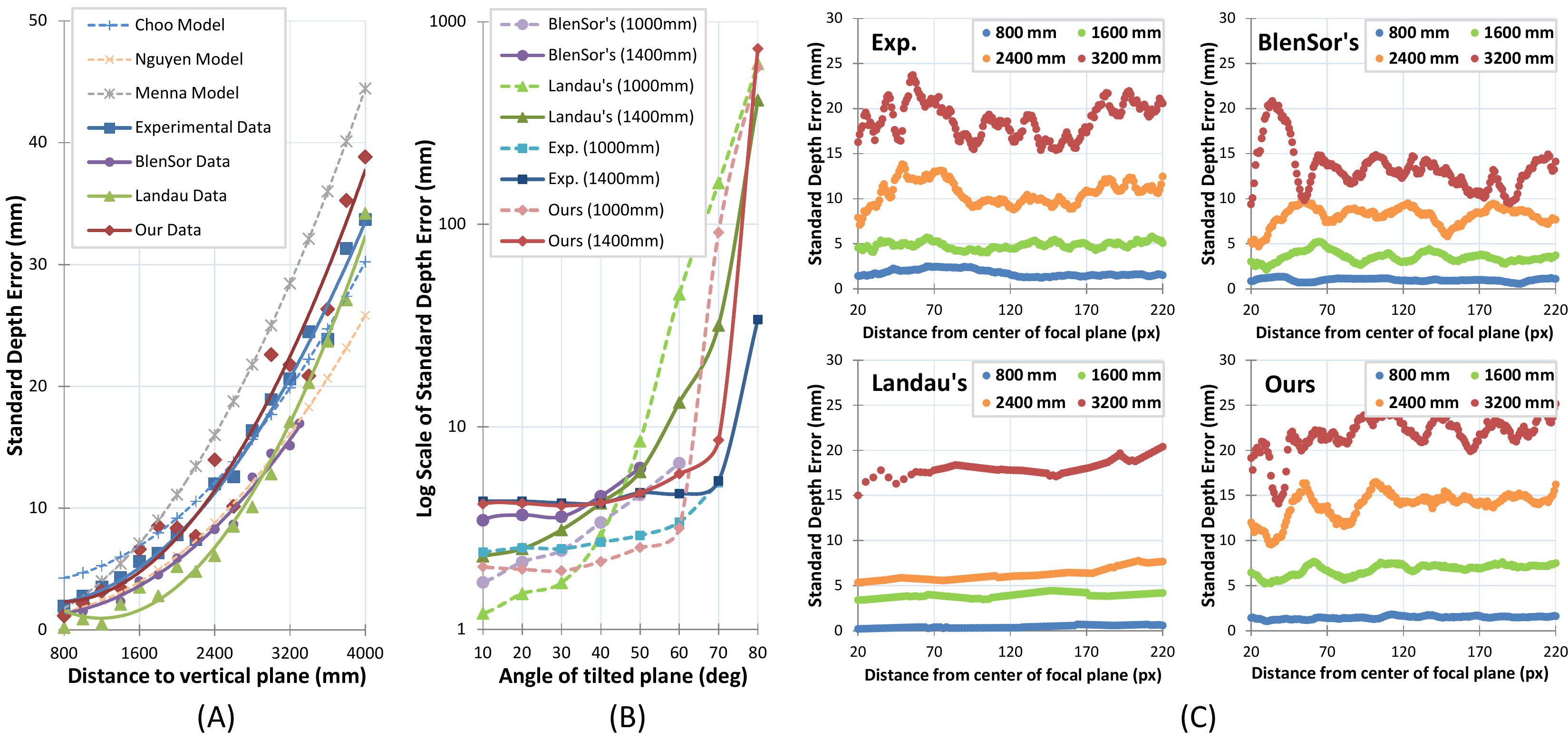}
  \caption{\textbf{Standard depth error (in mm)} as a function of (A) the distance (in mm) to a vertical flat wall for various fixed distances; (B) its tilt angle (in deg) for various fixed distances; (C) the radial distance (in px) to the focal center, plotted for the experimental images and the synthetic data from the various solutions, for various fixed distances.}
  \label{fig:error-dist}  
\end{figure*}

Most of the depth rendering tools chose to ignore background addition \eg by alpha compositing, causing significant discrepancy with real data and biasing the learner. Background modeling is hence another key component of \textit{DepthSynth}. Added backgrounds can be: (1) from static predefined geometry;
(2) from predefined geometry with motion; (3) with large amounts of random primitive shapes; (4) real captured scans (\eg from public datasets).

Optimized for GPU operations, the whole process can generate \textasciitilde 10 scans (VGA resolution) and their metadata (\eg viewpoints) per second on a middle-range computer (Intel E5-1620v2, 16GB RAM, NVidia Quadro K4200).

%
%
%

\section{Experiments and Results}
\label{sec:exp}

To demonstrate the accuracy and practicality of our method, we first analyze in Subsection~\ref{sec:exp-error} the depth error it induces when simulating the Kinect device, comparing with other simulation tools, experimental depth images and theoretical models for this device. 
In Subsection~\ref{sec:exp-est}, we adapt a state-of-the-art algorithm for classification and pose estimation to demonstrate how supervised 2.5D recognition methods benefit from using our data.
The pipeline developed for these evaluations makes use of Unity 3D Game Engine~\cite{engine9unity} (for rendering) and OpenCV~\cite{opencv_library} (for stereomatching).

\subsection{Depth Error Evaluation}
\label{sec:exp-error}

To validate the correctness of our simulation pipeline, we first replicate the set of experiments used by Landau~\etal~\cite{landau2015simulating,landau2016}, to compare the depth error induced by \textit{DepthSynth} to experimental values, as well as to the results from Landau~\etal ~\cite{landau2015simulating,landau2016}, from \textit{BlenSor}~\cite{gschwandtner2011blensor} and from 3 Kinect error models---respectively from Menna~\etal~\cite{menna2011geometric}, Nguyen~\etal~\cite{nguyen2012modeling} and Choo~\etal~\cite{choo2014statistical,landau2016}. 
All datasets consist of scans of a flat surface placed in front of the sensor at various distances and tilt angles to the focal plane. The experimental data was kindly provided by Landau~\cite{landau2015simulating,landau2016}.

Figure~\ref{fig:error-dist}(A) shows how the distance between the plane and the sensor influences the standard depth error in the resulting scans. The trend in our synthetic data matches well the one observed in experimental scans, and Choo~\etal model recalibrated by Landau on the same experimental data~\cite{landau2016}. As noted in~\cite{landau2016}, these models are based on experimental results which are inherently correlated to the characteristics of their environment and sensor. We could expect other data not to perfectly align with such models (as proved by the 
discrepancies among them). We can still notice that our synthetic images' quality degenerates slightly more for larger distances than the real scans, though our method behaves overall more realistically than the others.

In Figure~\ref{fig:error-dist}(B), we evaluate how the synthetic methods fares when gradually tilting the plane from orthogonal to almost parallel to the optical axis. The errors induced by our tool matches closely the experimental results for tilt angles below $70^{\circ}$, with some overestimation for steeper angles but a similar trend, unlike the other methods. It should be noted that for such incident angles, both real scans and \textit{DepthSynth} ones have most of the depth information missing, due to the poor reflection and stretching of the projected pattern(s), heavily impairing the reconstruction.

As a final experiment related to the error modeling, we compute the standard depth error as a function of the radial distance to the focal center. Again, Figure~\ref{fig:error-dist}(C) shows us that our pipeline behaves the most realistically, despite inducing slightly more noise for larger distances and thus more distorted pattern(s). \textit{DepthSynth} even satisfyingly reproduces the oscillating evolution of the noise when increasing the distance and reaching the edges of the scans---a well-documented phenomenon caused by "wiggling" and distortion of the pattern(s)~\cite{fursattel2016comparative,lachat2015first}. 

\subsection{Application to 6-DOF Pose Estimation and Classification}
\label{sec:exp-est}
{
Among the applications which can benefit from our pipeline, we formulate a 6-DOF camera pose recognition and classification problem from a single 2.5D scan into an image retrieval problem, supposing no real images can be acquired for the training of the chosen method. However in possession of the 3D models, we discretize $N_p$ camera poses, generate the synthetic 2.5D image for each pose and each object using \textit{DepthSynth}, and encode each picture via a discriminative, low-dimension image representation with its corresponding class and camera pose. We build this way a database for pose and class retrieval problems. Given an unseen image, its  representation is thus computed the same way and queried in the database to find the K-nearest neighbor(s) and return the corresponding class and pose.

To demonstrate the advantages of using \textit{DepthSynth} data irrespective of the selected features, we adapt Wohlhart~\etal ``triplet method''~\cite{Wohlhart15} which uses case-specific \textit{computer-crafted} image representations generated by a CNN. We thus use a CNN (LeNet structure~\cite{Lecun98} with custom hyper-parameters -- two $5\times5$ convolution layers, each followed by a ReLu layer and a $2\times2$ Max pooling layer; and finally two fully connected layers leading to the output one, also fully connected, as shown in Figure~\ref{fig:cnn_dl}) to learn the discriminating features by enforcing a loss function presented in~\cite{Wohlhart15}, over all the CNN weights ${\bf w}$:
\begin{equation}
  L = L_{triplet} + L_{pairwise} + \lambda{\|{\bf w}\|_2^2},
\end{equation}
where $L_{triplet}$ is the triplet loss function, $L_{pairwise}$ the pairwise one, and  $\lambda$ the regularization factor. A triplet is defined as ($p_{i}$, $p_{i}^{+}$, $p_{i}^{-}$), with $p_{i}$ one class and pose sampling point, $p_{i}^{+}$ a point \textit{close} to $p_{i}$ (similar class and/or pose) and $p_i^{-}$ another one \textit{far} from $p_i^+$ (different class and/or pose). A pair is defined as ($p_{i}$, $p'_i$), with $p_i$ one sampling point and $p'_i$ its perturbation in terms of pose and noise, to enforce proximity between the descriptors of similar data. Given a margin $m$, $L_{pairwise}$ is defined as the sum of the squared Euclidean distances between $f(p_{i})$ and $f(p'_i)$, and  $L_{triplet}$ as:
\begin{equation}
L_{triplet} = \sum_{\mathclap{(p_{i}, p_{i}^{+}, p_{i}^{-})}}max(0,1-\frac{\|{f(p_i)-f(p_i^{-})}\|_2}{\|{f(p_i)-f(p_i^{+})}\|_2+m}),
\end{equation}

Given such a state-of-the-art recognition method, we present the experiments to validate our solution and discuss their results.

\begin{figure}
  \centering
  \includegraphics[width=.95\linewidth]{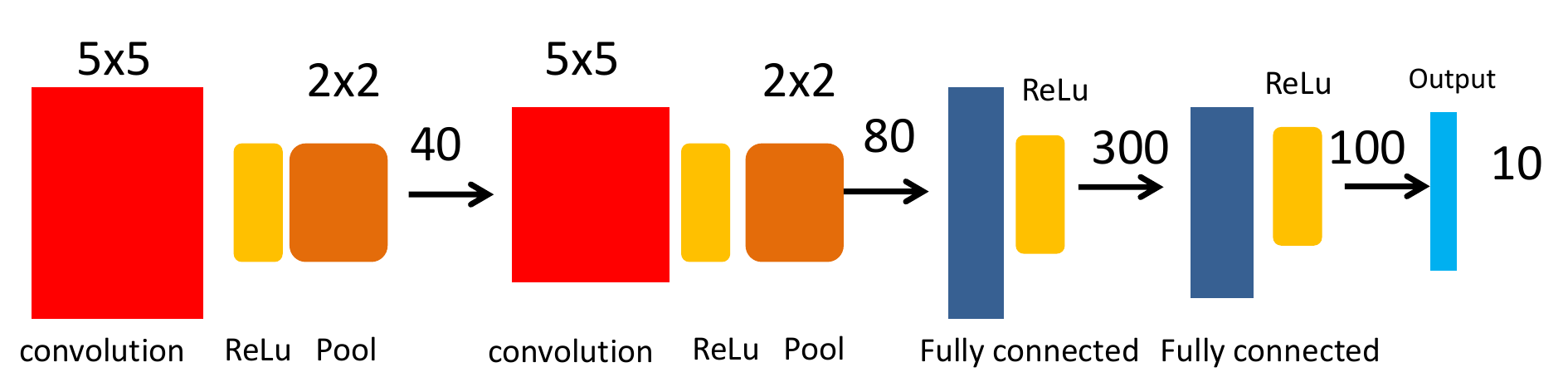}
  \caption{CNN architecture used in our experiment.}
  \label{fig:cnn_dl}  
\end{figure}

\begin{figure}[t]
  \centering
  \includegraphics[width=.95\linewidth]{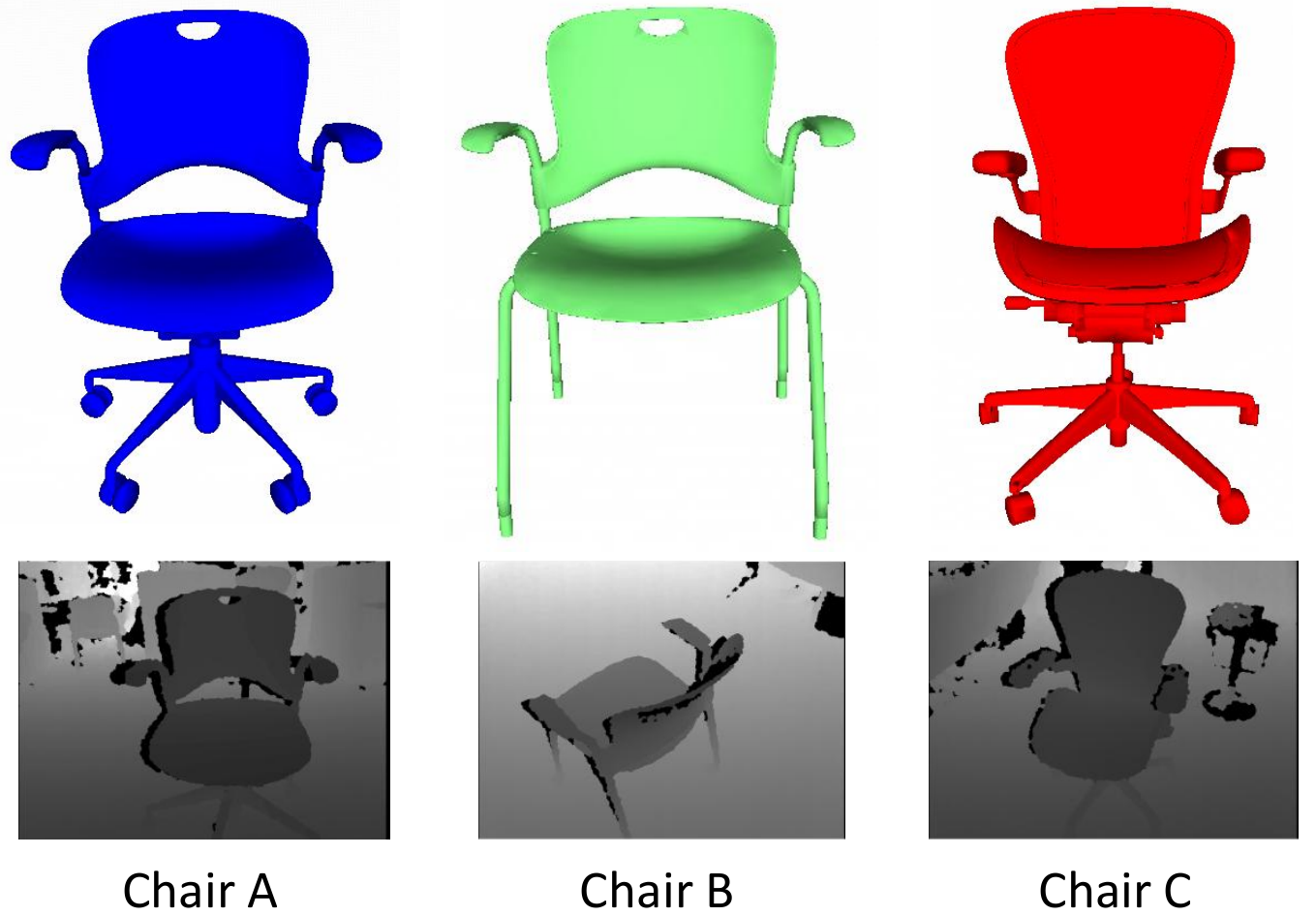}
  \caption{\textbf{CAD models and sample real images used in the experiments} (note the strong similarities among these chairs).}
  \label{fig:3_chairs}  
\end{figure}

\begin{figure}[t]
  \centering
  \includegraphics[width=1\linewidth]{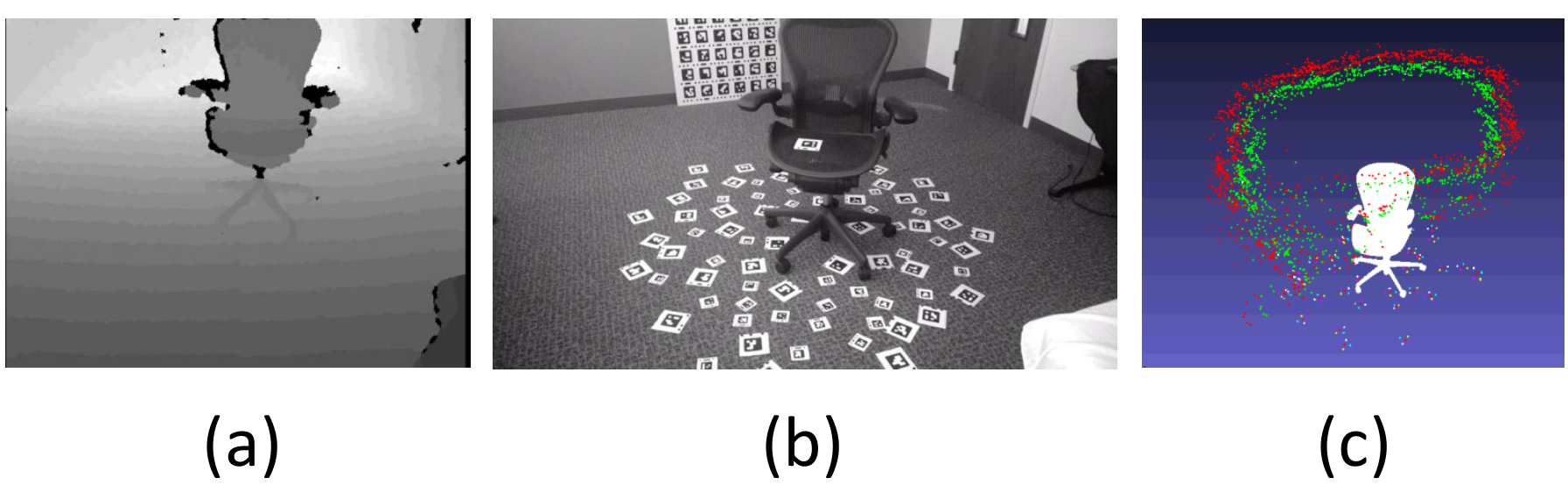}
  \caption{\textbf{Real data acquisition and processing.} (a) Sample real depth data. (b) Sample RGB data with markers. (c) Recovered trajectory (poses) for each samples, along markers for Chair C.}
  \label{fig:chair_exp}  
\end{figure}

\begin{figure}[t!]
  \centering
  \includegraphics[width=1\linewidth]{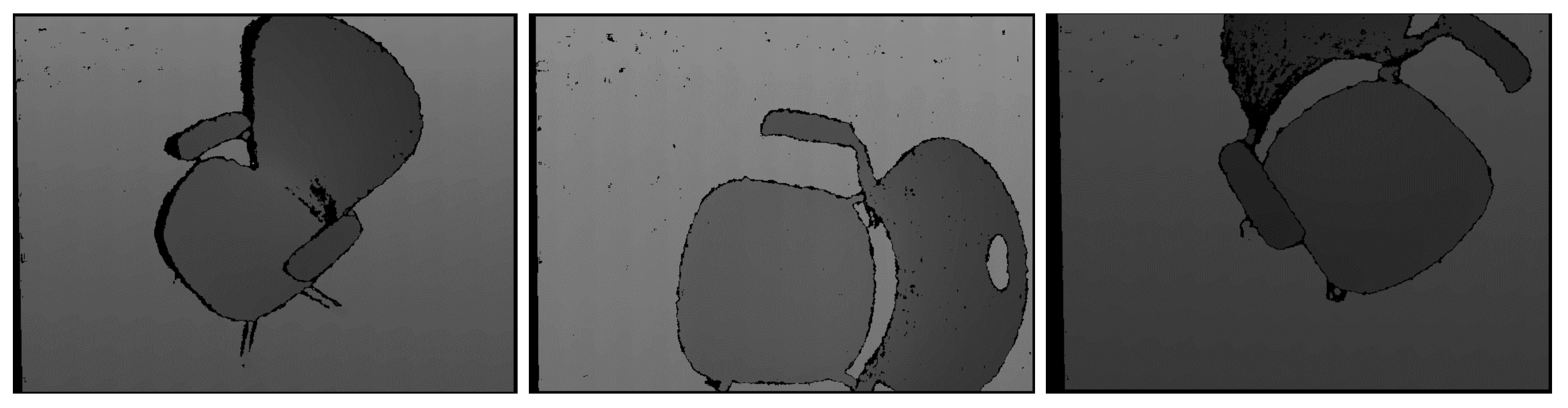}
  \caption{\textbf{Sample synthetic data generated by \textit{DepthSynth}.} }
  \label{fig:chair_exp_data}  
\end{figure}

\begin{figure*}[t!]
  \centering
  \includegraphics[width=1\linewidth]{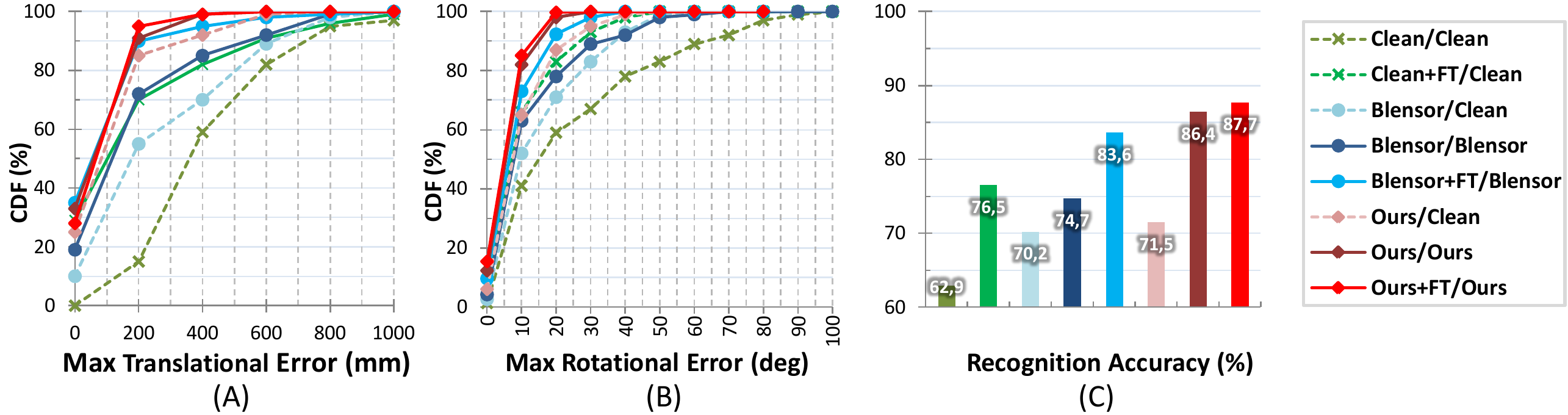}
  \caption{\textbf{Cumulative Distribution Function (CDF) on errors (A) in translation and (B) in rotation for pose estimation} on the \emph{Chair C} dataset. (C) \textbf{Classification results} over the \emph{3-Chairs} dataset, using the method trained over different datasets. (``FT'' = ``fine-tuning'')}
\label{fig:classify_res}
\end{figure*} 

\par\noindent
\textbf{Data Preparation} As target models for the experiment, we select three similar-looking office chairs, with their CAD models obtained from the manufacturers' websites (Figure~\ref{fig:3_chairs}). 
The following procedure is performed to capture the real 2.5D dataset and its ground-truth pose annotations: AR markers are placed on the floor around each chair, an Occipital Structure sensor is mounted on a tablet, and its infrared camera is calibrated according to the RGB  camera of the tablet. Using this table, an operator captures sequences of RGBD frames walking around the chairs (Figure~\ref{fig:chair_exp}).

In a comprehensive and redundant annotation procedure using robust Direct Linear Transform, we manually generate 2D-3D correspondences on chairs regions based on visual landmarks, choosing a representative set of approximately $60$ frames. These estimated camera poses and the detected 2D locations of markers are used to generate triangulated 3D markers locations in the CAD coordinate system. Given the objects' movable parts, the actual chairs deviate from their model. We thus iteratively reduce the deviation for the final ground-truth sequence by verifying the reprojections, and the consistency of the triangulated markers positions relative to the chairs elements.

The IR and RGB camera calibration parameters are then used to align the depth scans into the common 3D coordinate system. In a final fine-tuning step, the poses and 3D models are fed into the simulation pipeline, to generate the corresponding noiseless depth maps, used by an Iterative Closest Point method~\cite{besl1992method} to be aligned to the real images; optimizing the ground-truth for our real test dataset. 



\par\noindent
\textbf{Evaluation on Pose Estimation} As a first experiment, we limit the aforementioned approach to pose estimation only, training and testing it over the data of Chair C. For the CNN training, 30k synthetic depth scans rendered with CAD model and floor plane as shown in Figure~\ref{fig:chair_exp_data}, are used to form 100k samples (triplets + pairs). The learned representation is then applied for the indexation of all the 30k images, using FLANN~\cite{flann_pami_2014}. For testing, 
the representations of the 1024 depth images forming the real testing dataset are extracted and indexed. For each, the nearest neighbor's pose is then rendered and aligned to the input scan to refine the final 3D pose estimation.  

To demonstrate how the quality of the synthetic training data impacts the estimation, three different datasets are generated; resp. using noiseless rendering, \textit{BlenSor}, and \textit{DepthSynth}. Each dataset is used either for both representation-learning and indexing; or only for learning, with the clean dataset used for indexing. We also further apply some fine-tuning (FT) to the CNN training, feeding it with 200 real scans, forming 3k samples (triplets + pairs).
Estimated 3D poses are compared to the ground-truth ones, and the Cumulative Distribution Functions on errors in rotation and translation are shown in Figure~\ref{fig:classify_res}(A-B).

It reveals how the method trained over \textit{DepthSynth} data gives consistently better results on both translation and rotation estimations; furthermore not gaining much in accuracy after fine-tuning with real data.
%
%
%

\par\noindent
\textbf{Evaluation on Classification}  We consider in a second time the classification problem for the 3 chairs. Using the same synthetic training datasets extended to all 3 objects, we evaluate the accuracy of the recognition method over a testing dataset of 1024 real depth images for each chair, taking as final estimation the class of the nearest neighbor in the database for each extracted image representation.

Despite the strong similarities among the objects, the recognition method is performing quite well, as shown in Figure~\ref{fig:classify_res}(C). Again, it can be seen that it gives consistently better results when trained over our synthetic data; and that unlike other training datasets, ours doesn't gain much from the addition of real data, validating its inherent realism.

\section{Conclusion} 
\label{sec:cnc}
We presented \textit{DepthSynth}, a pipeline to generate large depth image datasets from 3D models, simulating the mechanisms of a wide panel of depth sensors to achieve unique realism with minimum effort. We not only demonstrated the improvements in terms of noise quality compared to state-of-the-art methods; but also went further than these previous works by showcasing how our solution can be used to train recent 2.5D recognition methods, outperforming the original results using lower-quality training data.

We thus believe this concept will prove itself greatly useful to the community, leveraging the parallel efforts to gather detailed 3D datasets. The generation of realistic depth data and corresponding 
ground truth can promote a large number of data-driven algorithms, by providing the training and benchmarking resources they need. 
We plan to further demonstrate this in a near future, applying our pipeline to tasks of larger-scale (e.g. semantic segmentation of the NYU depth dataset~\cite{eccv12:issi}, using SUNCG models~\cite{song2016semantic} as input for our pipeline) . We are also curious to compare---and maybe combine---\textit{DepthSynth} with recent GAN-based methods such those developed by Shrivastava~\etal~\cite{shrivastava2016learning} or Bousmalis~\etal~\cite{bousmalis2016unsupervised}.


{\small
\bibliographystyle{ieee}
\bibliography{eccv16_autoid}

\begin{thebibliography}{10}\itemsep=-1pt

\bibitem{aldoma2012point}
A.~Aldoma, Z.-C. Marton, F.~Tombari, W.~Wohlkinger, C.~Potthast, B.~Zeisl,
  R.~B. Rusu, S.~Gedikli, and M.~Vincze.
\newblock Point cloud library.
\newblock {\em IEEE Robotics \& Automation Magazine}, 1070(9932/12), 2012.

\bibitem{besl1992method}
P.~Besl and D.~McKay.
\newblock Method for registration of 3-d shapes.
\newblock In {\em Robotics-DL tentative}, pages 586--606. International Society
  for Optics and Photonics, 1992.

\bibitem{bousmalis2016unsupervised}
K.~Bousmalis, N.~Silberman, D.~Dohan, D.~Erhan, and D.~Krishnan.
\newblock Unsupervised pixel-level domain adaptation with generative
  adversarial networks.
\newblock {\em arXiv preprint arXiv:1612.05424}, 2016.

\bibitem{opencv_library}
G.~Bradski.
\newblock {\em Dr. Dobb's Journal of Software Tools}.

\bibitem{ijcai79:mbvs}
R.~A. Brooks, R.~Creiner, and T.~O. Binford.
\newblock The acronym model-based vision system.
\newblock In {\em Proceedings of the 6th International Joint Conference on
  Artificial Intelligence - Volume 1}, IJCAI'79, pages 105--113. Morgan
  Kaufmann Publishers Inc., 1979.

\bibitem{carlucci2016deep}
F.~M. Carlucci, P.~Russo, and B.~Caputo.
\newblock A deep representation for depth images from synthetic data.
\newblock {\em arXiv preprint arXiv:1609.09713}, 2016.

\bibitem{shapenet2015}
A.~X. Chang, T.~Funkhouser, L.~Guibas, P.~Hanrahan, Q.~Huang, Z.~Li,
  S.~Savarese, M.~Savva, S.~Song, H.~Su, J.~Xiao, L.~Yi, and F.~Yu.
\newblock {ShapeNet: An Information-Rich 3D Model Repository}.
\newblock Technical Report arXiv:1512.03012 [cs.GR], Stanford University ---
  Princeton University --- Toyota Technological Institute at Chicago, 2015.

\bibitem{cgf03:vsr}
D.-Y. Chen, X.-P. Tian, Y.-T. Shen, and M.~Ouhyoung.
\newblock On visual similarity based 3d model retrieval.
\newblock In {\em Computer graphics forum}, volume~22, pages 223--232. Wiley
  Online Library, 2003.

\bibitem{choo2014statistical}
B.~Choo, M.~Landau, M.~DeVore, and P.~A. Beling.
\newblock Statistical analysis-based error models for the microsoft kinecttm
  depth sensor.
\newblock {\em Sensors}, 14(9):17430--17450, 2014.

\bibitem{deng2009imagenet}
J.~Deng, W.~Dong, R.~Socher, L.-J. Li, K.~Li, and L.~Fei-Fei.
\newblock Imagenet: A large-scale hierarchical image database.
\newblock In {\em Computer Vision and Pattern Recognition, 2009. CVPR 2009.
  IEEE Conference on}, pages 248--255. IEEE, 2009.

\bibitem{engine9unity}
U.~G. Engine.
\newblock Unity game engine-official site.
\newblock {\em Online][Cited: October 9, 2008.] http://unity3d. com}, pages
  1534--4320.

\bibitem{fallon2012efficient}
M.~F. Fallon, H.~Johannsson, and J.~J. Leonard.
\newblock Point cloud simulation \& applications, 2012.
\newblock
  \href{http://www.pointclouds.org/assets/icra2012/localization.pdf}{http://www.pointclouds.org/assets/icra2012/localization.pdf}.
  Accessed: 2015-09-23.

\bibitem{nips12:d3d}
S.~Fidler, S.~Dickinson, and R.~Urtasun.
\newblock 3d object detection and viewpoint estimation with a deformable 3d
  cuboid model.
\newblock In {\em NIPS}, pages 611--619, 2012.

\bibitem{fursattel2016comparative}
P.~F{\"u}rsattel, S.~Placht, M.~Balda, C.~Schaller, H.~Hofmann, A.~Maier, and
  C.~Riess.
\newblock A comparative error analysis of current time-of-flight sensors.
\newblock {\em IEEE Transactions on Computational Imaging}, 2(1):27--41, 2016.

\bibitem{accv10:els}
A.~Geiger, M.~Roser, and R.~Urtasun.
\newblock Efficient large-scale stereo matching.
\newblock In {\em ACCV}, pages 25--38, 2011.

\bibitem{pr97:sr}
S.~Gold, A.~Rangarajan, C.~ping Lu, and E.~Mjolsness.
\newblock New algorithms for 2d and 3d point matching: Pose estimation and
  correspondence.
\newblock {\em Pattern Recognition}, 31:957--964, 1997.

\bibitem{gschwandtner2011blensor}
M.~Gschwandtner, R.~Kwitt, A.~Uhl, and W.~Pree.
\newblock Blensor: blender sensor simulation toolbox.
\newblock In {\em Advances in Visual Computing}, pages 199--208. Springer,
  2011.

\bibitem{sundatabase}
K.~E. A.~O. J.~Xiao, J.~Hays and A.~Torralba.
\newblock Sun database: Large-scale scene recognition from abbey to zoo.
\newblock In {\em Computer Vision and Pattern Recognition, 2010. CVPR 2010.
  IEEE Conference on}, pages 3485--3492. Springer, 2010.

\bibitem{keller2009real}
M.~Keller and A.~Kolb.
\newblock Real-time simulation of time-of-flight sensors.
\newblock {\em Simulation Modelling Practice and Theory}, 17(5):967--978, 2009.

\bibitem{konolige1998small}
K.~Konolige.
\newblock Small vision systems: Hardware and implementation.
\newblock In {\em Robotics Research}, pages 203--212. Springer, 1998.

\bibitem{iccv05:sls}
K.~N. Kutulakos and E.~Steger.
\newblock A theory of refractive and specular 3d shape by light-path
  triangulation.
\newblock In {\em IEEE ICCV}, pages 1448--1455, 2005.

\bibitem{lachat2015first}
E.~Lachat, H.~Macher, M.~Mittet, T.~Landes, and P.~Grussenmeyer.
\newblock First experiences with kinect v2 sensor for close range 3d modelling.
\newblock {\em The International Archives of Photogrammetry, Remote Sensing and
  Spatial Information Sciences}, 40(5):93, 2015.

\bibitem{icra11:lmvd}
K.~Lai, L.~Bo, X.~Ren, and D.~Fox.
\newblock A large-scale hierarchical multi-view rgb-d object dataset.
\newblock In {\em IEEE ICRA}, pages 1817--1824. IEEE, 2011.

\bibitem{landau2016}
M.~J. Landau.
\newblock {\em Optimal 6D Object Pose Estimation with Commodity Depth Sensors}.
\newblock PhD thesis, University of Virginia, 2016.
\newblock
  \href{http://search.lib.virginia.edu/catalog/hq37vn57m}{http://search.lib.virginia.edu/catalog/hq37vn57m}.
  Accessed: 2016-10-20.

\bibitem{landau2015simulating}
M.~J. Landau, B.~Y. Choo, and P.~A. Beling.
\newblock Simulating kinect infrared and depth images.
\newblock 2015.

\bibitem{Lecun98}
Y.~Lecun, L.~Bottou, Y.~Bengio, and P.~Haffner.
\newblock Gradient-based learning applied to document recognition.

\bibitem{iccv13:fpe}
J.~J. Lim, H.~Pirsiavash, and A.~Torralba.
\newblock Parsing ikea objects: Fine pose estimation.
\newblock In {\em IEEE ICCV}, pages 2992--2999. IEEE, 2013.

\bibitem{puc14:cgar}
Y.~Ma, K.~Boos, J.~Ferguson, D.~Patterson, and K.~Jonaitis.
\newblock Collaborative geometry-aware augmented reality with depth sensors.
\newblock In {\em Proceedings of the 2014 ACM International Joint Conference on
  Pervasive and Ubiquitous Computing: Adjunct Publication}, pages 251--254.
  ACM, 2014.

\bibitem{trs78:rep}
D.~Marr and H.~K. Nishihara.
\newblock Representation and recognition of the spatial organization of
  three-dimensional shapes.
\newblock {\em Proceedings of the Royal Society of London B: Biological
  Sciences}, 200(1140):269--294, 1978.

\bibitem{iros15:vn}
D.~Maturana and S.~Scherer.
\newblock Voxnet: A 3d convolutional neural network for real-time object
  recognition.
\newblock In {\em IEEE IROS}, September 2015.

\bibitem{menna2011geometric}
F.~Menna, F.~Remondino, R.~Battisti, and E.~Nocerino.
\newblock Geometric investigation of a gaming active device.
\newblock In {\em SPIE Optical Metrology}, pages 80850G--80850G. International
  Society for Optics and Photonics, 2011.

\bibitem{flann_pami_2014}
M.~Muja and D.~G. Lowe.
\newblock Scalable nearest neighbor algorithms for high dimensional data.
\newblock {\em Pattern Analysis and Machine Intelligence, IEEE Transactions
  on}, 36, 2014.

\bibitem{ismar11:kf}
R.~A. Newcombe, S.~Izadi, O.~Hilliges, D.~Molyneaux, D.~Kim, A.~J. Davison,
  P.~Kohli, J.~Shotton, S.~Hodges, and A.~Fitzgibbon.
\newblock Kinectfusion: Real-time dense surface mapping and tracking.
\newblock In {\em Proceedings of the 2011 10th IEEE International Symposium on
  Mixed and Augmented Reality}, ISMAR '11, pages 127--136, 2011.

\bibitem{nguyen2012modeling}
C.~V. Nguyen, S.~Izadi, and D.~Lovell.
\newblock Modeling kinect sensor noise for improved 3d reconstruction and
  tracking.
\newblock In {\em 3D Imaging, Modeling, Processing, Visualization and
  Transmission (3DIMPVT), 2012 Second International Conference on}, pages
  524--530. IEEE, 2012.

\bibitem{peters2008bistatic}
V.~Peters and O.~Loffeld.
\newblock A bistatic simulation approach for a high-resolution 3d pmd (photonic
  mixer device)-camera.
\newblock {\em International Journal of Intelligent Systems Technologies and
  Applications}, 5(3-4):414--424, 2008.

\bibitem{pharr2016physically}
M.~Pharr, W.~Jakob, and G.~Humphreys.
\newblock {\em Physically based rendering: From theory to implementation}.
\newblock Morgan Kaufmann, 2016.

\bibitem{rematas2014image}
K.~Rematas, T.~Ritschel, M.~Fritz, and T.~Tuytelaars.
\newblock Image-based synthesis and re-synthesis of viewpoints guided by 3d
  models.
\newblock In {\em Computer Vision and Pattern Recognition (CVPR), 2014 IEEE
  Conference on}, pages 3898--3905. IEEE, 2014.

\bibitem{cviu15:synt}
A.~Rozantsev, V.~Lepetit, and P.~Fua.
\newblock {On rendering synthetic images for training an object detector}.
\newblock {\em Computer Vision and Image Understanding}, 2015.

\bibitem{icra09:l3d}
A.~Saxena, J.~Driemeyer, and A.~Y. Ng.
\newblock Learning 3-d object orientation from images.
\newblock In {\em IEEE ICRA}, pages 4266--4272, 2009.

\bibitem{schlick1994inexpensive}
C.~Schlick.
\newblock An inexpensive brdf model for physically-based rendering.
\newblock In {\em Computer graphics forum}, volume~13, pages 233--246. Wiley
  Online Library, 1994.

\bibitem{cvpr11:hpr}
J.~Shotton, A.~Fitzgibbon, M.~Cook, T.~Sharp, M.~Finocchio, R.~Moore,
  A.~Kipman, and A.~Blake.
\newblock Real-time human pose recognition in parts from a single depth image.
\newblock In {\em IEEE CVPR}, June 2011.

\bibitem{shrivastava2016learning}
A.~Shrivastava, T.~Pfister, O.~Tuzel, J.~Susskind, W.~Wang, and R.~Webb.
\newblock Learning from simulated and unsupervised images through adversarial
  training.
\newblock {\em arXiv preprint arXiv:1612.07828}, 2016.

\bibitem{eccv12:issi}
N.~Silberman, D.~Hoiem, P.~Kohli, and R.~Fergus.
\newblock Indoor segmentation and support inference from rgbd images.
\newblock In {\em ECCV}, pages 746--760. Springer, 2012.

\bibitem{icra14:bb}
A.~Singh, J.~Sha, K.~S. Narayan, T.~Achim, and P.~Abbeel.
\newblock Bigbird: A large-scale 3d database of object instances.
\newblock In {\em IEEE ICRA}, pages 509--516. IEEE, 2014.

\bibitem{nips12:crdl}
R.~Socher, B.~Huval, B.~Bath, C.~D. Manning, and A.~Y. Ng.
\newblock Convolutional-recursive deep learning for 3d object classification.
\newblock In {\em NIPS}, pages 665--673, 2012.

\bibitem{cvpr15:sund}
S.~Song, S.~P. Lichtenberg, and J.~Xiao.
\newblock Sun rgb-d: A rgb-d scene understanding benchmark suite.
\newblock In {\em IEEE CVPR}, pages 567--576, 2015.

\bibitem{song2016semantic}
S.~Song, F.~Yu, A.~Zeng, A.~X. Chang, M.~Savva, and T.~Funkhouser.
\newblock Semantic scene completion from a single depth image.
\newblock {\em arXiv preprint arXiv:1611.08974}, 2016.

\bibitem{bmvc10:3dcad}
M.~Stark, M.~Goesele, and B.~Schiele.
\newblock Back to the future: Learning shape models from 3d cad data.
\newblock In {\em BMVC}, pages 106.1--106.11. BMVA Press, 2010.

\bibitem{iccv15:mvcnn}
H.~Su, S.~Maji, E.~Kalogerakis, and E.~G. Learned{-}Miller.
\newblock Multi-view convolutional neural networks for 3d shape recognition.
\newblock In {\em IEEE ICCV}, 2015.

\bibitem{ax15:rcnn}
H.~Su, C.~R. Qi, Y.~Li, and L.~J. Guibas.
\newblock Render for {CNN:} viewpoint estimation in images using cnns trained
  with rendered 3d model views.
\newblock {\em CoRR}, abs/1505.05641, 2015.

\bibitem{eccv14:rtf}
Y.~Wang, J.~Feng, Z.~Wu, J.~Wang, and S.-F. Chang.
\newblock From low-cost depth sensors to cad: Cross-domain 3d shape retrieval
  via regression tree fields.
\newblock In {\em ECCV}, September 2014.

\bibitem{Wohlhart15}
P.~Wohlhart and V.~Lepetit.
\newblock Learning descriptors for object recognition and 3d pose estimation.
\newblock In {\em IEEE CVPR}, pages 3109--3118, 2015.

\bibitem{cvpr15:3dsp}
Z.~Wu, S.~Song, A.~Khosla, F.~Yu, L.~Zhang, X.~Tang, and J.~Xiao.
\newblock 3d shapenets: A deep representation for volumetric shapes.
\newblock In {\em IEEE CVPR}, pages 1912--1920, 2015.

\bibitem{tpami99:sfs}
R.~Zhang, P.-S. Tsai, J.~Cryer, and M.~Shah.
\newblock Shape-from-shading: a survey.
\newblock {\em IEEE TPAMI}, 21(8):690--706, 1999.

\end{thebibliography}
}

\end{document}